# Deep Learning Based Automatic Video Annotation Tool for Self-Driving Car

**N.S.Manikandan,**
TIFAC-CORE in Automotive Infotronics,
Vellore Institute of Technology,
Katpadi, Vellore, Tamilnadu, India-632014
nsmanikandan@vit.ac.in

**K.Ganesan,**
TIFAC-CORE in Automotive Infotronics,
Vellore Institute of Technology,
Katpadi, Vellore, Tamilnadu, India-632014
kganesan@vit.ac.in

Abstract

In a self-driving car, objection detection, object classification, lane detection and object tracking are considered to be the crucial modules. In recent times, using the real time video one wants to narrate the scene captured by the camera fitted in our vehicle. To effectively implement this task, deep learning techniques and automatic video annotation tools are widely used. In the present paper, we compare the various techniques that are available for each module and choose the best algorithm among them by using appropriate metrics. For object detection, YOLO and Retinanet-50 are considered and the best one is chosen based on mean Average Precision (mAP). For object classification, we consider VGG-19 and Resnet-50 and select the best algorithm based on low error rate and good accuracy. For lane detection, Udacity's 'Finding Lane Line' and deep learning based LaneNet algorithms are compared and the best one that can accurately identify the given lane is chosen for implementation. As far as object tracking is concerned, we compare Udacity's 'Object Detection and Tracking' algorithm and deep learning based Deep Sort algorithm. Based on the accuracy of tracking the same object in many frames and predicting the movement of objects, the best algorithm is chosen. Our automatic video annotation tool is found to be 83% accurate when compared with a human annotator. We considered a video with 530 frames each of resolution 1035 x 1800 pixels. At an average each frame had about 15 objects. Our annotation tool consumed 43 minutes in a CPU based system and 2.58 minutes in a mid-level GPU based system to process all four modules. But the same video took nearly 3060 minutes for one human annotator to narrate the scene in the given video. Thus we claim that our proposed automatic video annotation tool is reasonably fast (about 1200 times in a GPU system) and accurate.

Introduction:

The world health organization has surveyed 180 countries in the world and has reported that 1.25 million people die every year due to road accidents. The death rate is high in low-income countries [1]. One of the ways we can reduce this death rate is to use driverless cars. In a recent survey, 69% of respondents have reported that driverless cars are safer than human driven cars [2]. In a driverless car, the safety critical control functions such as steering, acceleration and breaking happens without any driver's interference [3]. There are many levels of automation of driverless cars. One of the levels is to use computer vision. The object detection and classification are the crucial problems in computer vision. The detection and classification systems detect and classify the various objects that are present on the road, especially vehicles, pedestrians, and stationary objects on the road side such as traffic signals, sign boards, light poles. For the development of detection and classification models, real time training datasets are needed. But these datasets use a large quantity of images. The objects present in these dataset are annotated manually by human beings. During the annotation process, they draw bounding boxes around the identified objects and also narrate (store) the properties of these objects. The annotators generally use the open-source (manual) annotation tools [4].

Using these tools one can create bounding boxes for object localization, draw polygons for object segmentation and add labels using text against the chosen regions. The annotated data are stored in many formats such as text XML, JSON, YOLO[5], ILSVRC[6] etc. Manual annotation is not only expensive but also a time consuming process. For example the object detection database of ILSVRC [6] needs about 42 seconds to draw a bounding box around an object [7]. To make this process of bounding box annotation to be cheaper and accurate, two different strategies are adopted by researchers. They are semi-automatic and fully-automatic annotation methods.

During the development of semi-automatic annotation tool Dim P. Papadopoulos et. al[8]. have reduced the annotation time by using a center-click annotation architecture and in their tool they asked the annotators to click at the center of the object present in an image. This method was found to be fast and also reduced the annotation time by 9X to 18X times. Adithya Subramanian et al[9]. have presented a new methodology for quickly annotating the data using click-supervision and hierarchical object detection techniques. They used semi-automatic method. The task of annotations was split between the human and a neural network. The proposed framework by them was 3-4 times faster than the standard manual annotation methods.

Dim P. Papadopoulos et. al[10]. have proposed a full-automatic annotation tool wherein the objects were detected using a learning algorithm. The human annotators were simply verifying the bounding boxes. Their method reduced the annotation time by about 6X-9X times. . Zhujun Xiao et.al[11]. have designed a self-annotating image generation tool by combining camera with passive wireless localizer. The pedestrian and vehicle detection modules were used as examples. They have demonstrated the feasibility, benefits, and challenges of an automatic image annotation system.

mopedist, motor cyclist, cyclist, other-two-wheeler and non-descript. The detected two wheeler object is characterized further by ten properties such as occlusion, head occlusion, feet occlusion, direction, movement, lane assignment, rotation, pose, lighting and bounding box measurement (size). Table 3 classifies a detected human as pedestrian and non-descript. The detected pedestrian is further characterized by nine properties such as occlusion, head occlusion, feet occlusion, direction, movement, height, strange pose, lighting and bounding box measurement (size).

Table 1: Vehicle properties

| Object type | Occlusion | Bottom Occlusion | Direction | Movement | Lane assignment | Lane change detection | Rotation | Pose | Lighting | Size |
|---|---|---|---|---|---|---|---|---|---|---|
| car/ bus/ truck/ other-vehicle/ non-descript | none/ partial/ full | true/ false | preceding/ oncoming | moving/ stationary/ parked | unknown/-2/-1/0/+1/+2 | true/ false | relevant/ irrelevant | rear/ rearright/ rearleft/ front/ frontright/ frontleft/ side | normal/ unsharp/ glare | minx, miny, maxx, maxy |

Table 2: Two-Wheeler properties

| Object type | Occlusion | Head Occlusion | Feet occlusion | Direction | Movement | Lane assignment | Rotation | Pose | Lighting | Size |
|---|---|---|---|---|---|---|---|---|---|---|
| mopedist/ motorcyclist/ cyclist/ other-two-wheeler/ non-descript | none/ partial/ full | true/ false | true/ false | preceding/ oncoming | moving/ stationary/ parked | unknown/-2/-1/0/+1/+2 | relevant/ irrelevant | rear/ rearright/ rearleft/ front/ frontright/ frontleft/ side | normal/ unsharp/ glare | minx, miny, maxx, maxy |

Table 3: Pedestrian properties

| Object type | Occlusion | Head occlusion | Feet occlusion | Direction | Movement | Height | Feet occlusion | Strange Pose | Lighting | Size |
|---|---|---|---|---|---|---|---|---|---|---|
| pedestrian/ non-descript | none/ partial/ full | true/ false | true/ false | NN/NE/ NW/SS/ SE/SW/ EE/WW | moving/ stationary/ parked | adult/ child | true/ false | true/ false | normal/ unsharp/ glare | minx, miny, maxx, maxy |

For building the annotation for vehicle detection and classification, many automotive companies are using their own object detection techniques and their properties selection criteria. Some of the properties chosen and the object detection approach used are shown in Tables above. Table 1 classifies the detected vehicle as car, bus, truck, other-vehicle and non-descript. The detected vehicle is further characterized by using ten properties such as occlusion, bottom occlusion, direction, movement, lane assignment, lane change detection, rotation, pose, lighting and bounding box measurement (size). Table 2 classifies the detected two wheeler as

II. Proposed model

The various frames of a given video are fed as input data to our proposed model. The model is expected to find the objects such as vehicles, two-wheelers and pedestrians using object detection algorithm and extract it's properties such as occlusion, pose, direction etc. using object classification techniques. The module is supposed to detect the lanes on the fly using lane detection algorithm. By tracking all the objects using an appropriate algorithm, the model has to identify the movements and record whether they change the lanes. These steps are clearly shown in fig. 1. below The model is divided into four sub models. They are object detection, properties identification during classification, lane detection and tracking every detected object for movement and lane change detection. The details of the proposed model are described below.

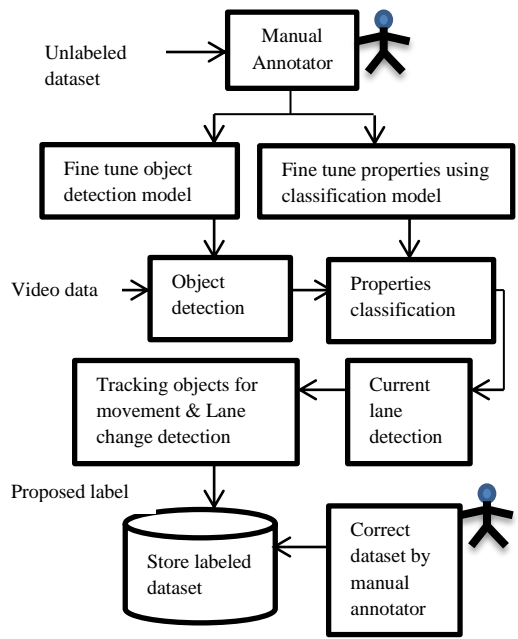

Fig. 1. Block diagram of proposed model

*A. Datasets:*

We use a custom dataset. The samples were collected from the Asian countries' road side videos. A normal webcam was installed on the top of the car. The roads were classified as highways and city roads. Some roads had the center median/ barrier and some roads didn't have. The input videos were manually annotated as vehicles, two wheeler and pedestrian along with their characteristics such as occlusion, bottom occlusion, head occlusion, direction, movement, pose, lane identification, lane change etc. The custom dataset collected manually had 28,450 annotated objects along with the properties. This was done using 29 annotators and has taken 4 days, each day they were working for about 8 hours (55680 minutes). Thus the average time taken to annotate each object was roughly about 2 minutes.

*B. Object detection:*

The first step of our automatic annotation tool is object detection. At this stage our tool detects the objects as vehicles/ two-wheelers/ pedestrians. If the object is detected as vehicle then the tool classifies the object as car/bus/truck/other-vehicle/non-descript. If the object is detected as a two-wheeler then the system needs to further classify it as mopedist /motor-cyclist /cyclist /other-two-wheeler /non-descript. The human is identifies as non-descript/pedestrian. Here the non-descript property is used for all three categories. We find whether the detected bounding box's height and width is less than 30 pixels. If so, we assign the property of that object as non-descript. In some scenarios the vehicles or two-wheelers use to come in the opposite direction but the road is divided by a median/ barrier. Then the vehicle or two-wheeler is classified as non-descript. There are many object detection algorithms used in deep learning such as R-CNN, Fast-R-CNN, Faster-R-CNN, YOLO, SSD, Retinanet etc. For our object detection step we use YOLO[4] and Retinanet-50[12] algorithms. The results obtained are compared and shown in our result analysis section. Figure 2a. below shows the object detected and figure 2b shows the segmentation of the detected objects. The segmented objects are then passed on to the classification module wherein properties of the objects are automatically identified apart from classifying them further.

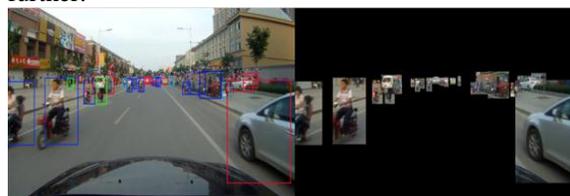

Fig 2: (a) Object detection        (b) segmentation of detected object

*C. Classification and properties assignment:*

The results obtained from the object detection module are passed onto the properties assignment and classification module. The Tables 1 to 3 explained before describe the various properties to be identified automatically for every object. If the object is a vehicle then the properties to be identified are: object type, occlusion, bottom occlusion, direction, movement, lane assignment, lane change detection, rotation, pose, lighting, and size. If the object detected is a two-wheeler then properties to be assigned are: object type occlusion, head occlusion, feet occlusion, direction, movement, lane assignment, rotation, pose, lighting, and size. If the object detected is a pedestrian then the properties to be assigned are: object type occlusion, head occlusion, feet occlusion, direction, movement, height, strange pose, lighting and size.  Here the object type and size for all the objects have been detected by using object detection algorithm. The lane assignment, lane change detection and movement are performed by lane detection and tracking algorithm as third and fourth modules.  The remaining properties for all objects are: occlusion, bottom occlusion, head occlusion, feet occlusion, direction, rotation, height, pose, strange pose and lighting. They are all detected using classification algorithm. Here the occlusion means how much the object is occluded. It can be partial, full or none. The bottom occlusion means, the vehicle's bottom is not visible. The head and feet occlusion refers to how much occlusion, the two-wheeler rider/ pedestrian's head and feet has encountered. The direction refers to the vehicle/two-wheeler's direction. Using classification methods we can say whether the object is oncoming or preceding or at the side of our ego vehicle. But for the pedestrian, it is classified as one of the eight directions. The rotation depends on the direction. If the vehicle is oncoming or preceding then the rotation is relevant otherwise it is irrelevant. The height for a pedestrian is to classify whether the pedestrian is an adult or child. The pose for the vehicle/two-wheeler is to classify it based on which side of it is visible. That is front or back etc. But the strange pose for a pedestrian refers to cases where he/she may be carrying a baby or doing something on the road. The final property is about lighting wherein we classify the object's clarity. It can be clear or glared or un-sharp. The following figure 3 shows the possible directions of a pedestrian against the ego car. It shows eight directions of a pedestrian and one pedestrian moving along North West (NW) direction.

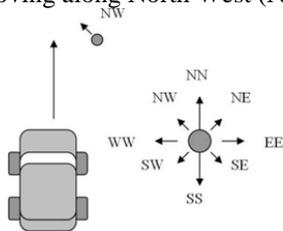

Fig 3: Pedestrian direction

To classify the properties of every object we have used two classification algorithms namely VGG-19[13] and Resnet-50[14]. The results are shown in the result analysis section. The following figs 4(a) to 4(f) show the properties of various object categories.

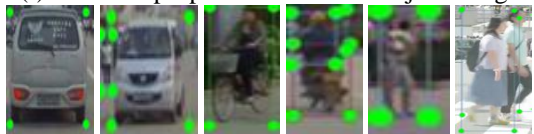

Fig 4 (a)    (b)    (c)    (d)    (e)    (f)

Figure 4a shows the object type as car and its properties are: occlusion: none, bottom occlusion : false, direction: preceding, rotation: relevant, pose: back, and lightning: normal. Figure 4b shows the object type as car, and its properties are occlusion: none, bottom occlusion : false, direction: oncoming, rotation: relevant, pose: front left, and lighting: normal. The figure 4c shows the object type as cyclist, and its properties are: occlusion: none, head occlusion: false, feet occlusion: false, direction: oncoming, rotation: relevant, pose: front left and lighting: normal. Figure 4d has an object type: mopedist, and its properties are: occlusion: none, head occlusion: false, feet occlusion: false, direction: preceding, rotation: relevant, pose: back left and lighting: normal. In figure 4e the object type is pedestrian, and its properties are: occlusion: none, head occlusion: false, feet occlusion: false, direction: WW, height: adult, strange pose: true and lighting: normal. In figure 4f the object type is pedestrian, and its properties are: occlusion: partial, head occlusion: false, feet occlusion: true, direction: WW, height: adult, strange pose: false and lighting: normal. Our next module is lane detection.

*D. Lane detection*

In our lane detection module we have considered up to 6 lanes. They are labeled as unknown, -2, -1, 0, 1, 2. Here 0 refers to the lane in which our ego vehicle is moving, -1 refers to the lane on the immediate left side, -2 refers to two lanes (farthest) on the left side, 1 refers to the lane on the immediate right side of our ego vehicle and 2 refers to two lanes (farthest) on the right side of the ego vehicle. But unknown refers to the target object which is not in the path of the ego vehicle. For example, the target may be parked on the parking area or pedestrian is standing on the foot path or the target vehicle is moving in the opposite side of the road where the road is divided by a median/ barrier. We have chosen Udacity's self driving car lane detection algorithm[15] and the LaneNet[16] deep learning algorithm. The results are compared and are given in the result analysis section. The output of LaneNet is shown in figure 5a, in which the instance segmentation of the lane is shown. The four lane marks detected on the road are coloured with pink, blue, green and yellow line colors. Here some lines were not properly segmented. To overcome this problem Hough Transform was used to draw the straight line over the segmented lane and it is clearly shown in figure 5b.

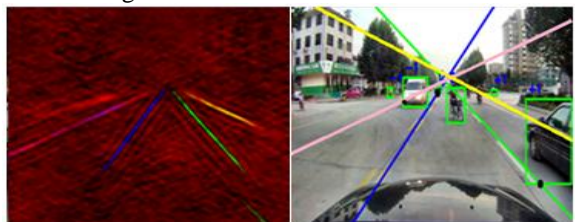

Fig 5: (a) Instance lane segmentation (b)Detected lane with lane assignment

After the lane detection, we take up the lane assignment task. The lane in which our ego vehicle is moving is assigned the lane number 0. In Figure 5b it lies between the blue and green line labeled lanes. The lane lying between pink and blue lines is assigned the lane number -1. It is at the left side of the ego vehicle. The lane on the left most side of pink line is assigned the lane number -2. The lane lying between green and yellow is assigned lane number +1. It is at the right hand side of our ego vehicle. The lane on the right most side of ego vehicle is assigned the lane number +2. Any object that is not belonging to any lanes is given the unknown lane assignment. Here the problem is, if the preceding vehicle moves on the green line then how can we assign whether the vehicle is moving in lane number 0 or +1. For this problem, one can take the bottom line of the detected object and make an intersection with the underlying color line labeled lane. Based on the intersection point one can find whether the maximum portion of the bottom line lies on the left or to the right side of the lane and accordingly lane number is assigned. For example, in the above figure 5b the bottom line of the detected car which is preceding near the ego vehicle intersects with the green color lane and its maximum portion fall in to the +1 lane. So the lane assignment for this car is chosen as lane number +1.

*E. Tracking*

This is the last step of our fully-automatic annotation tool. In the first step, vehicles, two-wheelers and pedestrians were detected by using object detection algorithm. The detected objects were passed on to our classification algorithm to find out the properties of each object like oncoming, partial occlusion for vehicles or pedestrian moving in the direction of north (NN) and carrying a baby as strange pose etc. The third step was to detect the lane in which the vehicle is moving. The results are then passed to this tracking step to identify whether the detected vehicles or two wheelers are moving or parked or stationary. Using the movement property selection one can find whether the detected vehicle or two-wheeler is changing its lane using lane change identification property. A unique ID is created for each object in every frame. For tracking we choose Udacity's self-driving car vehicle detection and tracking algorithm [17] and the Deep SORT[18] multi object tracking deep learning algorithm. The results are reported in the result analysis section. In the tracking algorithm, each detected object is given a tracking ID. This ID is used as a unique ID of the object and is stored along with the properties of the objects. The last property we need to find is movement. The aim here is to find whether the object is moving or parked or stationary. For this movement detection inputs are object bounding boxes with tracked ID and lane number assigned. The bounding box of the detected object in the previous frame and the bounding box of the detected object in the current frame are checked to see whether they possess same tracking ID. If so, it is passed to ORB (Oriented FAST and Rotated BRIEF) [19] feature extraction technique which provides key points from the detected object's image. Key points of an image are unique features extracted from the image. Key points obtained from previous and current frame for every object are passed to BF Matcher (Brute Force Matching)[19] algorithm to find the matching point from both object images. From the matching points one can get the distance between the two image coordinates and is called as the pixel distance of two ROIs (Region Of Interests). One can collect all matching point's distance and find its mean distance. If the mean distance is greater than six pixel distance then that object is assigned to moving class. If the pixel distance is less than six pixels then the object is still in any one of the lanes (-2,-1,0,1,2). Then the object belongs to stationary class. If the pixel distance is less than six pixels but the object is in the unknown lane then the object is assigned to parking class. Figure 6 below shows the movement detection of detected object.

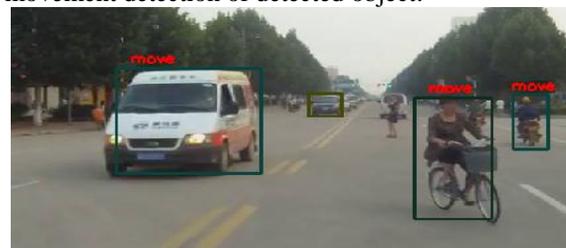

Fig 6: Object tracking and movement property detection

III Result analysis:

In this section we analyze the results obtained using our fully-automatic annotation tool. Here we choose the best method for our final algorithm.

*A. Object detection:*

For object detection the metrics used for the comparative study between YOLO and Retinanet is training loss, Average Precision (AP), and mean AP (mAP)[20]. For testing a video with 530 frames each of resolution 1035 x 1800 pixels was used. At an average each frame had about 15 objects. We manually annotated (7950 objects) and utilized them for this experiment. The Table 4 below shows the Average Precision (AP) of various objects and compared it with YOLO and RetinaNet-50 algorithms. YOLO's mAP is 34.35, whereas Retinanet-50's mAP is 48.3. The Table 5 below shows the Accuracy (%) of various objects detected and compared it with YOLO and RetinaNet-50 algorithms. From table 5 the over-all (9 class of objects) mean accuracy of YOLO is 60.12% and that of Retinanet-50 is 82.13%. Figure 9a shows YOLO's training loss and figure 9b shows the Retinanet-50's training loss. The Retinanet-50 has lower loss than YOLO. Hence we choose the Retinanet-50 algorithm as our object detection algorithm.

### B. Properties classification

From the output of our object detection algorithm namely Retinanet-50, the detected objects are passed on to the properties classification algorithm. Here we notice the lack of accuracy in the object detection algorithm. Figure 7 shows an object in which the occlusion is classified as none during the manual annotation but the automatic classification algorithm has declared it as partial occlusion due to the fact that the object detection algorithm has not properly drawn the bounding box. Table 6 shows the comparative study between VGG-19 and Resnet-50 for both vehicle occlusion and pedestrian direction properties. The figure 9c and 9e show the training loss and validation loss of VGG-19 and Resnet-50 for vehicle occlusion and pedestrian direction properties. Resnet-50 has lower loss for both training and validation. Figure 9d and 9f shows the training average and validation average of VGG-19 and Resnet-50 for vehicle occlusion and pedestrian direction properties. The Resnet-50 has the highest average during training and validation phase. The Resnet-50 has given high accuracy (overall 97%) than VGG-19 model. Hence the ResNet-50 algorithm is selected for properties classification.

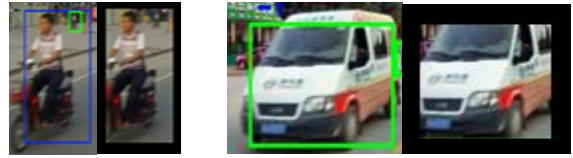
Fig 7: wrong classification due to wrong bounding box size

### C. Lane detection

The Udacity's lane detection algorithm was not able to detect the lanes for the Asian road datasets and is shown in figure 8a below. It detects the objects but was not able to properly detect the lane. Hence the LaneNet is chosen for the lane detection technique. The number of detected objects in the correct lane was 6678. Some of the objects are not detected by our proposed object detection algorithm and are shown in figure 8b as red circles (pedestrian and mopedist). Number of manually annotated objects for the video which has 530 frames is 7950. Thus LaneNet algorithm produced 84 % accuracy against the manual annotation.

### D. Object tracking

The metrics used for multi object tracking are Multi Object Tracking Accuracy (MOTA)[21], Multi Object Tracking Precision (MOTP)[21], Mostly Tracked (MT)[21] and Mostly Lost (ML)[21]. The comparison between the Udacity tracking algorithm and Deep SORT algorithm using the above four metrics are shown in Table 7.

Table 4: Average Precision of YOLO-V2 vs RetinaNet-50

|  | Car (AP) | Bus (AP) | Truck (AP) | Other-Vehicle (AP) | Mopedist (AP) | Motorcyclist (AP) | Cyclist (AP) | Other-Two-Wheeler (AP) | Pedestrian (AP) |
|---|---|---|---|---|---|---|---|---|---|
| YOLO | 38.6 | 35.2 | 40.0 | 30.9 | 34.6 | 33.3 | 31.1 | 29.8 | 35.7 |
| Retinanet-50 | 54.1 | 56.7 | 48.7 | 41.6 | 49.8 | 41.3 | 50.2 | 42.7 | 49.6 |

Table 5: Accuracy of YOLO-V2 vs RetinaNet-50

|  | Car (%) | Bus (%) | Truck (%) | Other-Vehicle (%) | Mopedist (%) | Motor-cyclist (%) | Cyclist (%) | Other-Two-Wheeler (%) | Pedestrian (%) |
|---|---|---|---|---|---|---|---|---|---|
| YOLO | 74 | 60 | 80 | 40 | 57.3 | 60.8 | 40 | 75 | 54 |
| Retinanet-50 | 86.9 | 87 | 93 | 81 | 82.9 | 71 | 80 | 69 | 88.4 |

Table 6: Accuracy of VGG-19 vs ResNet-50

|  | Vehicle occlusion | Pedestrian direction |
|---|---|---|
| VGG-19 | 86.6 | 79.5 |
| ResNet-50 | 97.45 | 96.8 |

Table 7: Metrics for Tracking (Udacity project vs Deep SORT)

|  | MOTA | MOTP | MT | ML |
|---|---|---|---|---|
| Udacity tracking | 15.9 | 69.8 | 6.4% | 47.9% |
| Deep SORT | 71.4 | 79.1 | 34% | 18.2% |

Table 8: Processing time in CPU vs GPU

|  | CPU system | GPU system |
|---|---|---|
| Object detection | 13.2 minutes | 51 seconds |
| Object classification | 12.5 minutes | 44 seconds |
| Lane identification | 7.5 minutes | 22 seconds |
| Object tracking | 10.5 minutes | 37 seconds |

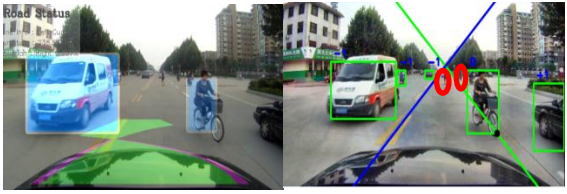

Fig 8 (a) Udacity lane detection    (b) LaneNet lane detection

The accuracy of multi object tracking (MOTA) using deep SORT is 71.4%. Hence the Deep SORT is chosen as the object tracking algorithm for our fully-automatic annotation tool development.

*E. Processing Time:*

The time taken to complete the various modules for the test video (530 frames) are shown in Table 8 above. We have tested our video using the CPU based system whose specification is Intel i5 processor, 16GB RAM, and Ubuntu 16.04 operating system and the specification of our GPU based system is Intel XEON processor, NVIDIA Quadro P1000 graphics card, 32GB RAM and Ubuntu 16.04 operating system. The input video with 530 frames takes 43 minutes to complete the automatic annotation process in the CPU system whereas the GPU system has taken 2.58 minutes only. The manual annotation by a single annotator has taken 3060 minutes. Thus we claim that our proposed automatic annotation tool is reasonably fast (about 1200 times when compared with a GPU system) and also accurate one.

IV Conclusion:

We have briefly described few algorithms that are used for object detection, classification, lane identification and object tracking models. We have analyzed the algorithms and chosen the best algorithm in each model. The chosen algorithms were used in our fully-automatic annotation tool to create the automatic training dataset for the self-driving car. Our results (based on the test video with 530 frames with each frame containing about 15 objects) shows that the Retinanet-50 is the best algorithm for object detection model as it had an accuracy of 82%. The ResNet-50 is chosen for properties classification algorithm which provides 97% accuracy. The LaneNet is chosen as the best algorithm for lane identification whose accuracy is 84%. For object tracking Deep SORT algorithm is chosen as the best algorithm with 71% accuracy. The average accuracy of our fully-automatic annotation tool is 83% which is 17% lesser than the manual annotation. Our fully-automatic annotation tool supports manual annotation for corrections. And the overall processing time using the CPU system is 43 minutes and 2.58 minutes in the GPU system. The manual annotation took about 3060 minutes for one annotator. So our fully-automatic annotation tool which uses GPU system is about 1200 times faster than the manual annotator.

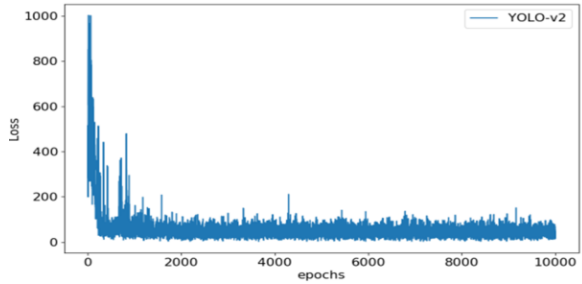

Fig 9. (a) YOLO-V2 Training Loss

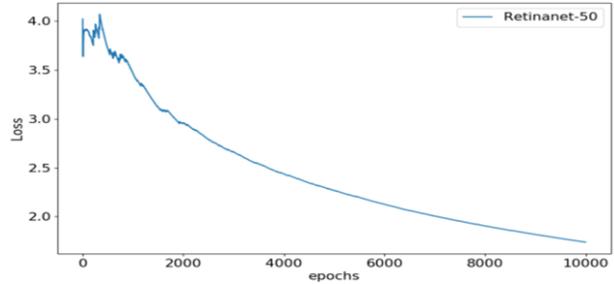

Fig 9. (b) RetinaNet-50 Training Loss

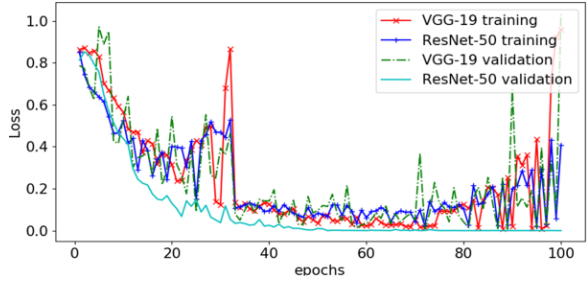

Fig 9. (c) Loss of Vehicle Occlusion

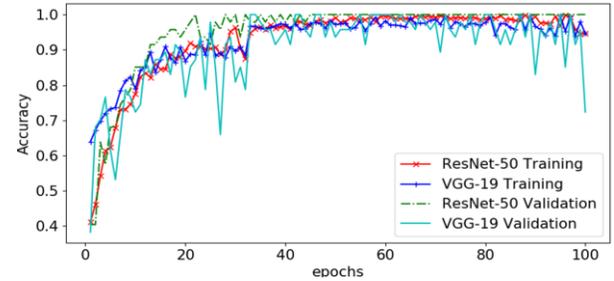

Fig 9. (d) Accuracy of Vehicle Occlusion

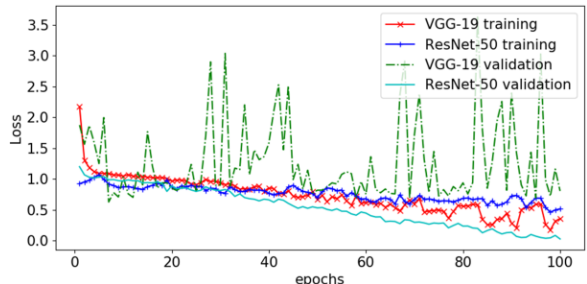

Fig 9. (e) Loss of Pedestrian direction

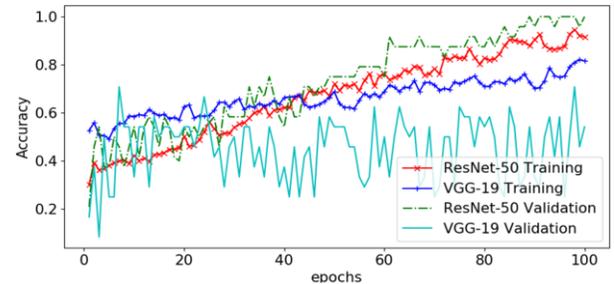

Fig 9. (f) Accuracy of Pedestrian direction

**ACKNOWLEDGEMENT:**
The proposed work was carried out at the TIFAC-CORE in Automotive Infotronics Research Centre at VIT, Vellore. We would like to thank DST, Government of India for the support given to the centre.